\documentclass[transmag]{IEEEtran}
\usepackage{algorithm}
\usepackage{multirow}
\usepackage{makecell}
\usepackage{amsfonts}
\usepackage{booktabs}
\usepackage{color}

\usepackage{subfigure}

\usepackage{times}
\usepackage{epsfig}
\usepackage[cmex10]{amsmath}
\usepackage{amssymb}
\usepackage{multirow}
\usepackage{stfloats}
\usepackage{tabularx}
\usepackage{pifont}
\usepackage{makecell}
\usepackage{threeparttable}
\usepackage{mathtools}
\usepackage{xcolor}
\usepackage{graphicx}
\usepackage{caption}
\usepackage{algorithmic}
\usepackage{url}
\begin{document}
%\title{Large Scale Image Classification and Retrieval with Sparse Product Quantization}
\title{Attribute-aware Pedestrian Detection in a Crowd}

\author{
Jialiang~Zhang, Lixiang~Lin, Yang~Li, Yun-chen Chen, Jianke~Zhu~\IEEEmembership{Senior~Member,~IEEE},\\ Yao~Hu, Steven~C.H.~Hoi~\IEEEmembership{Fellow,~IEEE}
\IEEEcompsocitemizethanks{
\IEEEcompsocthanksitem Jialiang~Zhang, Lixiang~Lin, Yang~Li, and Jianke Zhu are with the College of Computer Science, Zhejiang University, Hangzhou, China, 310027.\protect\\
E-mail: \{zjialiang, lxlin liyang89, jkzhu\}@zju.edu.cn.

\IEEEcompsocthanksitem Yun-chen~Chen and Steven C.H. Hoi are with School of Information Systems, Singapore Management University, Singapore.\protect\\
E-mail: \{jeanchen, chhoi\}@smu.edu.sg

\IEEEcompsocthanksitem Yao~Hu is with company of Alibaba, Youku Cognitive and Intelligent Lab.\protect\\
E-mail: yaoohu@alibaba-inc.com
\IEEEcompsocthanksitem Jianke Zhu is the Corresponding Author.
}
\thanks{}}
\maketitle

\begin{abstract}
    Pedestrian detection is an initial step to perform outdoor scene analysis, which plays an essential role in many real-world applications. Although having enjoyed the merits of deep learning frameworks from the generic object detectors, pedestrian detection is still a very challenging task due to heavy occlusion and highly crowded group. Generally, the conventional detectors are unable to differentiate individuals from each other effectively under such a dense environment. To tackle this critical problem, we propose an attribute-aware pedestrian detector to explicitly model people's semantic attributes in a high-level feature detection fashion. Besides the typical semantic features, center position, target's scale and offset, we introduce a pedestrian-oriented attribute feature to encode the high-level semantic differences among the crowd. Moreover,  a novel attribute-feature-based Non-Maximum Suppression~(NMS) is proposed to distinguish the person from a highly overlapped group by adaptively rejecting the false-positive results in a very crowd settings. Furthermore, a novel ground truth target is designed to alleviate the difficulties caused by the attribute configuration and extremely class imbalance issue during training. Finally, we evaluate our proposed attribute-aware pedestrian detector on two benchmark datasets including CityPersons and CrowdHuman. The experimental results show that our approach outperforms state-of-the-art methods at a large margin on pedestrian detection.
    
\end{abstract}


% \fi
\IEEEpeerreviewmaketitle

\section{Introduction}

Pedestrian detection is an initial step to perform outdoor scene analysis, which plays an essential role in many real-world applications, such as autonomous vehicles, security surveillance, robotics and so on.

Being a special case of generic object detection, pedestrian detectors~\cite{cai2016unified,8060595,DBLP:conf/cvpr/ZhangBS17} inherit a lot of successful techniques from the conventional anchor-based approaches, such as Faster RCNN~\cite{zhang2016faster} and SSD~\cite{liu2016ssd}. Most recently, anchor-free detectors like CornerNet~\cite{law2018cornernet} and CenterNet~\cite{zhou2019objects}, are emerging as the promising methods that are able to enjoy the merits of flexibility in network model design. To this end, we present a pedestrian detector under the anchor-free object detection framework.

Comparing to generic object detection, the pedestrian detector has its own characteristics and challenges in the real-world, such as vastly different scales, poor appearance conditions, and extremely severe occlusion in crowd scenarios. 

\begin{figure}[t]
    \begin{center}
        \includegraphics[width=1\linewidth]{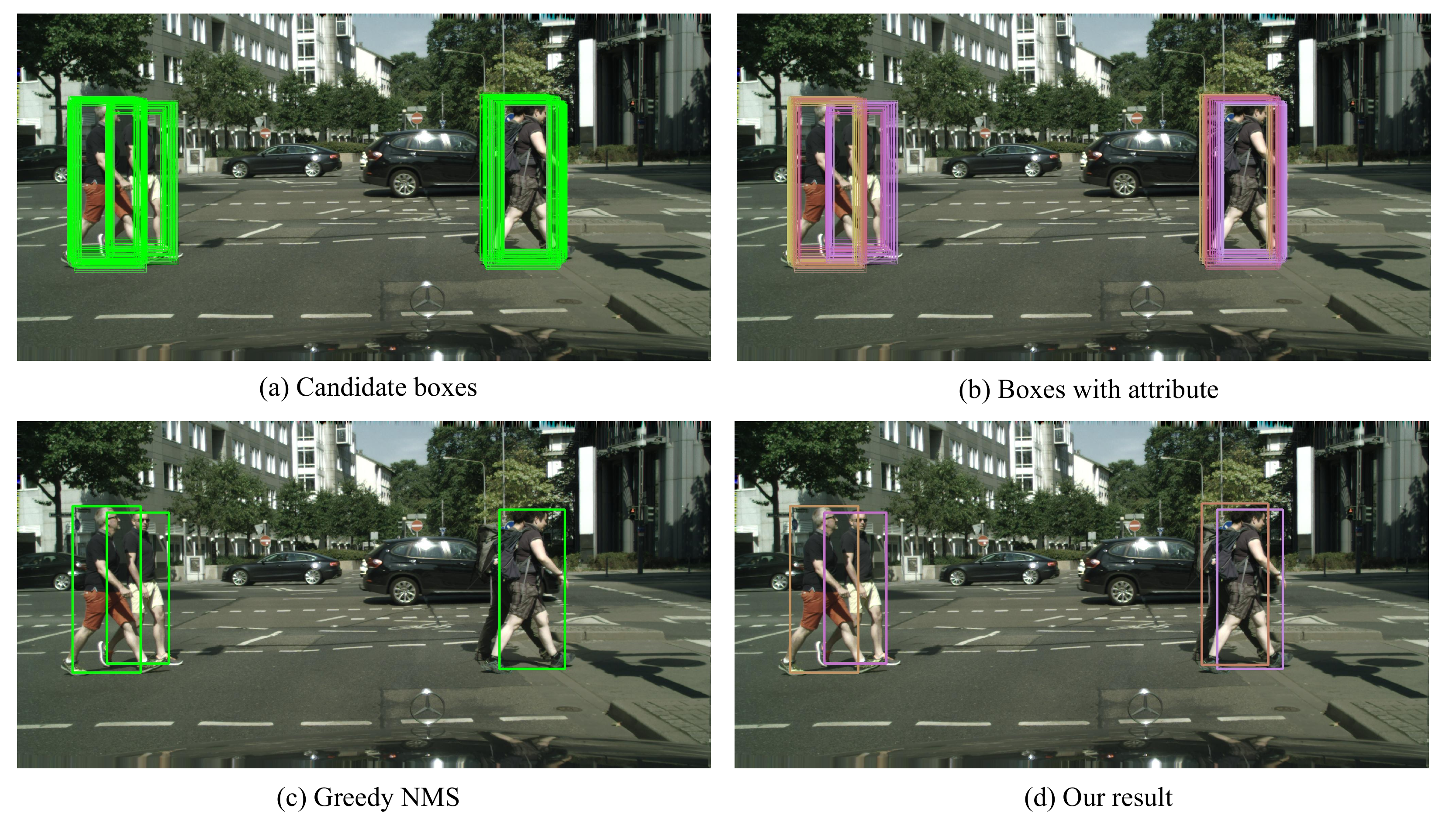}
    \end{center}
    %\vspace{-0.1in}
    \caption{Illustration of our attribute-aware pedestrian detector. (a) The predicted bounding boxes of detector~\cite{liu2019high} without NMS. (b) The bounding boxes using our method colored by a 3-D vector that is mapped from the attribute vector through embedding. (c) The result of detector~\cite{liu2019high} after greedy-NMS. (d) Our result after our proposed attribute-aware NMS.}
    \label{fig:ill}
    %\vspace{-0.1in}
\end{figure}

To localize a visual object, a modern detector usually produce a large amount of overlapped proposals around the target, and then employ the Non-Maximum Suppression~(NMS) to select the most possible bounding boxes based on likelihood scores. As the target usually stands without highly overlapped cases, this pipeline works well in the general object detection scenario. However, occlusions commonly occur in pedestrian detection. There are $48.8\%$ annotated pedestrians occluded by each other in the CityPersons dataset~\cite{DBLP:conf/cvpr/ZhangBS17}. In addition, Wang \textit{et al.}~\cite{wang2018repulsion} found that the level of crowd occlusion is even higher in the CrowdHuman dataset~\cite{shao2018crowdhuman}. Usually, it is hard for the conventional pipeline to chose the bounding boxes in a crowd setting. As shown in Fig.~\ref{fig:ill}, it is challenging to distinguish the bounding boxes generated by a single target or those from multiple people occluded together using a single filtering threshold in the greedy NMS. This is because a lower threshold may lead to missing highly overlapped objects while a higher one often has more false positives. 

To deal with pedestrian detection in a crowd, recent methods either design the proper regression losses for more compact boxes generation~\cite{wang2018repulsion, zhang2018occlusion} or develop the post-processing NMS technique. In fact, NMS is greatly affected by the feature extraction, and they should be dealt with in a whole pipeline.   

To address the above limitations, we propose a novel attribute-aware pedestrian detector by tackling the crowd occlusion issue from the original feature space side in this paper. Inspired by a human being's visual perception process, our algorithm tries to understand the crowd scene as a whole and encode high-level semantic information for each individual rather than alleviating ambiguities from the last separate component like NMS in the detector. To this end, we formulate the pedestrian detection problem as a feature detection task as in~\cite{liu2019high}, and explicitly model each people's semantic attribute as an embedding vector in a high-level feature space. Combing the basic semantic features including center position, target's scale and offset, our proposed attribute-aware feature encodes the semantic differences among the persons in a crowd as shown in Fig.~\ref{fig:ill}. Moreover, a density label and a diversity metric are generated as ground truth targets for the attribute-aware training stage. With the very rich semantic information of the proposed attribute feature, a novel attribute-aware NMS is introduced to adaptively reject the false-positive proposals and distinguish the individuals in a very crowd setting. 

In summary, the main contributions of this paper are: 1) a novel attribute map that converts each positive instance into a feature vector to encode both density and diversity information simultaneously; 2) an enhanced ground truth target to define attribute loss and address extremely class imbalance issue during training; 3) a novel attribute-aware NMS algorithm by considering the high-level semantic information of each predicted box to effectively refine the detection results; 4) experiments on two standard benchmark datasets~\cite{DBLP:conf/cvpr/ZhangBS17,shao2018crowdhuman} show that our proposed approach outperforms state-of-the-art pedestrian detectors at a large margin.

%------------------------------------------------------------------------
\section{Related Work}
As pedestrian detection can be viewed as the object localization for the special category, we firstly briefly review the related works on generic object detection. Then, we introduce the recent progress on it. Moreover, we will investigate the techniques related to solving the detection in a crowd, including occlusion handling and Non-Maximum Suppression.

\subsection{Generic object detection}
Early generic object detectors are based on the hand-crafted features~\cite{dollar2014fast,felzenszwalb2009object,papageorgiou2000trainable}, which aim at finding and classifying the candidates in a sliding window paradigm or region proposal strategy. With the development of deep learning, the modern detectors can be categorized into two classes, anchor-based methods, and anchor-free approaches. 

The anchor-based object detectors have dominated in past years. The two-stage detectors including Fast-RCNN~\cite{girshick2015fast}, Faster-RCNN~\cite{ren2015faster} and their variants~\cite{7792742} generate the object proposals in the first stage and classify them in the second stage. On the other hand, the single-stage methods like SSD~\cite{liu2016ssd}, YOLOv2~\cite{redmon2017yolo9000} manage to perform detection and classification simultaneously using the feature maps without a separate proposal generation stage.

Recently, anchor-free methods achieve promising results, which directly formulate the detection problem as a regression task having a simple network structure without anchor boxes.
CornerNet~\cite{law2018cornernet} treats object detection as keypoint detections and their associations, which are free from windows and anchors. CenterNet~\cite{zhou2019objects} views an object as a single point, which is the center point of its bounding box. Then, these center points regress to all other object properties, such as size, 3D location, orientation, and even pose. Our proposed method essentially belongs to the anchor-free detection framework.

\subsection{Pedestrian Detection}
Traditional pedestrian detectors~\cite{dollar2014fast,nam2014local,zhang2015filtered} rely on integral channel features with sliding window strategy to localize each target. CNN-based detectors are mainly based on Faster RCNN framework.
In~\cite{zhang2016faster}, RPN is employed to generate proposals and provide CNN features followed by a boosted decision forest. Zhang \textit{et al.}~\cite{DBLP:conf/cvpr/ZhangBS17} applies a plain Faster-RCNN for pedestrian detection. Cai \textit{et al.}~\cite{cai2016unified} employ multi-scale feature maps to match pedestrians of different scales.
Mao \textit{et al.}~\cite{mao2017can} use extra features to further improve performance.
Song \textit{et al.}~\cite{song2018small} propose to detect the object by predicting the top and bottom vertexes.
Liu \textit{et al.}~\cite{liu2018learning} present the asymptotic localization fitting strategy to gradually refine the localization results. Later, they suggest an anchor-free framework to detect center and scale in pedestrian detection~\cite{liu2019high}. 

\subsection{Occlusion Handling} 
Occlusions are quite common for pedestrian detection in a crowd. CityPersons~\cite{zhang2017citypersons} and CrowdHuman datasets~\cite{shao2018crowdhuman} are collected especially for crowd scenarios. Part-based models~\cite{enzweiler2010multi,mathias2013handling,ouyang2012discriminative,zhou2017multi,zhou2018bi,tian2015deep,8345752} are usually used to deal with occlusions. They learn a series of part detectors, and then design some mechanisms to fuse each result to localize partially occluded pedestrians. In~\cite{tian2015pedestrian}, attention mechanism is used to represent occlusion patterns. RepLoss~\cite{wang2018repulsion} and OR-CNN~\cite{zhang2018occlusion} design two novel regression losses to generate more compact boxes to make them less sensitive to the NMS threshold in crowded scenes.

\subsection{Non-Maximum Suppression}
NMS is a critical post-processing step for object detectors. Greedy-NMS suffers from suppressing objects in crowd scenarios. Instead of discarding predicted boxes during suppressing, soft-NMS~\cite{bodla2017soft} decreases the scores of neighbors by an increasing function of their overlap with the higher scored bounding box.
Learning NMS~\cite{hosang2017learning} trains a deep neural network to perform the NMS function using the predicted boxes and their corresponding scores. The relation module is employed to learn the NMS function in~\cite{hu2018relation} as an end-to-end general object detector. Tychsen \textit{et al.}~\cite{tychsen2018improving} and Jiang \textit{et al.}~\cite{jiang2018acquisition} learn extra localization confidences to guide a better NMS. Idrees \textit{et al.}~\cite{idrees2018composition} and Ranjan \textit{et al.}~\cite{ranjan2018iterative} estimate a crowd density map in people counting task. Rujikietgumjorn~\textit{et al.}~\cite{rujikietgumjorn2013optimized} propose a quadratic unconstrained binary optimization solution to suppress detection boxes, where a hard threshold is still applied to blindly suppress detection boxes as greedy-NMS. Lee \textit{et al.}~\cite{lee2016individualness} combine the determinantal point process with individual prediction scores to optimally select the final detection boxes. Adaptive-NMS~\cite{liu2019adaptive} proposes to estimate the density of predicted boxes to set an adaptive threshold. Despite promising results, its NMS threshold is merely determined by the density of the current box with the largest score, which ignores the relationship among boxes. Thus, selecting the bounding boxes is still the most critical issue for pedestrian detection.

%------------------------------------------------------------------------
\begin{figure*}[!t]
	\begin{center}
		\includegraphics[width=1.\linewidth]{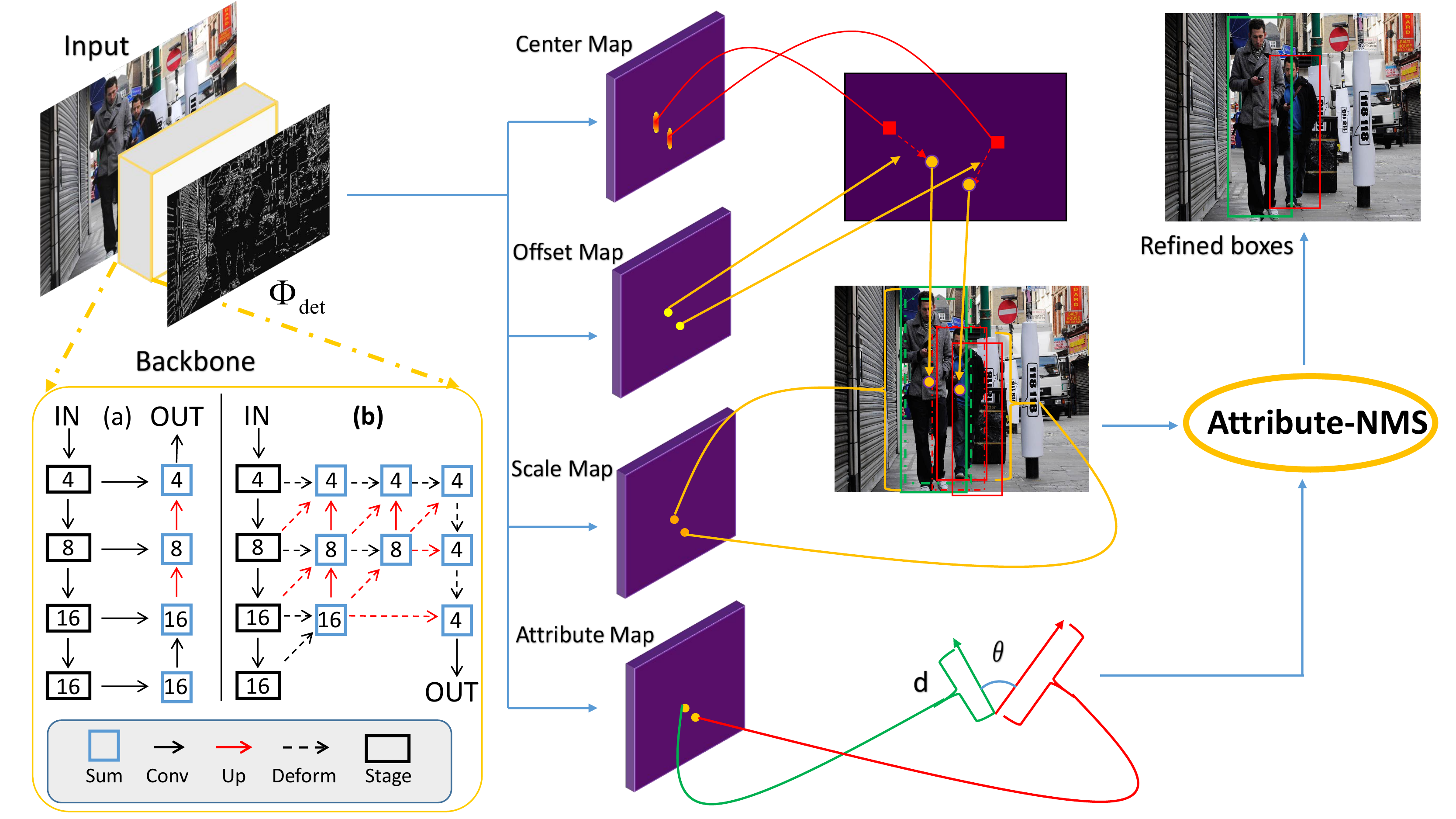}
	\end{center}
	%\vspace{-0.1in}
	\caption{ Overall architecture of our proposed detector. Detection is performed on $\Phi_{det}$ with the down-sampling rate $r=4$. Our detector consists of four detection heads. Attribute map encodes the density and diversity information of each positive instance into an embedding vector.
	The length $d$ of embedding vector denotes the density while the angle $\theta$ between two of them indicates their similarity. Attribute-aware NMS refines the predicted boxes by taking into account of both density and diversity information from attribute map. The bottom left is our model diagram for (a) ResNet-50 and (b) DLA-34.
	} \label{fig:arc}
	%\vspace{-0.1in}
\end{figure*}
\section{Attribute-aware Pedestrian Detection}

In this section, we first give the overall architecture of the proposed attribute-aware pedestrian detector. To explicitly distinguish a pedestrian from feature representation, four high-level semantic branches are proposed in our network structure. Then, we describe the details of our optimization targets design for conventional detection branches, center, scale, and offset maps, which consists of the definition of ground truth targets as well as the objective functions.
Moreover, a novel attribute loss is presented for training our proposed attribute map to encode both density and diversity information.
Finally, we introduce an attribute-aware NMS algorithm to fully exploit our proposed attribute-aware structure.

\subsection{Attribute-aware Pedestrian Detector}

Under the anchor-free detection framework~\cite{liu2019high}, we introduce four branches into our proposed detector to describe the pedestrian. Each branch produces a feature map to represent a property of the pedestrian in the image. The overall architecture is illustrated in Fig.~\ref{fig:arc}. The backbone is truncated from a standard network pre-trained on ImageNet~\cite{russakovsky2015imagenet}, such as ResNet-50~\cite{he2016deep} and deep layer aggregation network DLA-34~\cite{yu2018deep}.

\subsubsection{Feature Extraction}
There is a total of five stages in ResNet-50 and DLA-34 with the down-sampling rate $2, 4, 8, 16$ and $32$, respectively. We incorporate semantics from the last four feature maps, in which shallower feature maps provide more precise localization information while deeper feature maps contain more semantic features. Specifically, we augment ResNet-50 with a standard FPN structure~\cite{lin2017feature}, where an up-sampling network allows for a higher-resolution output and skip connections for richer semantics. DLA-34 is an image classification network that employs iterative deep hierarchical aggregation to enhance the feature representation. We augment the skip connections with deformable convolution from lower layers to the output as CenterNet~\cite{zhou2019objects}. The detailed model diagrams are depicted in Fig.~\ref{fig:arc}. For the last stage, the dilated convolution is adopted to keep the spatial feature size as $1/16$ of the input image. Final detection is performed on an aggregated feature map $\Phi_{det}$ with the down-sampling rate $r=4$.

\subsubsection{Detection Heads}
We introduce four detection heads, which predict four high-level semantic features, including center, scale, offset and attribute map, respectively. All four predicted feature maps are with the same size as $\Phi_{det}$ that is $H/r \times W/r$, where input image $I \in \mathcal{R}^{W \times H \times 3}$. Each of them is appended after the aggregated feature maps $\Phi_{det}$. Specifically, we attach one $3 \times 3$ convolution layer with $256$ channels to $\Phi_{det}$. A final $1 \times 1$ convolution is appended to produce the desired output.

The center map aims at localizing the central position of positive instances while the scale map predicts the height and width of them. 
The offset map is to recover the quantization error caused by the down-sampling rate $r$ for defining the location of positive instances. 
The attribute map is designed for converting each positive instance into an embedding vector which encodes both diversity and density information simultaneously. We fuse the center map, scale map, and offset map to produce the predicted bounding boxes. These boxes and attribute map are further fed into attribute-aware NMS algorithm in order to effectively produce the refined boxes.

\begin{figure}[!t]
	\begin{center}
		\includegraphics[width=1\linewidth]{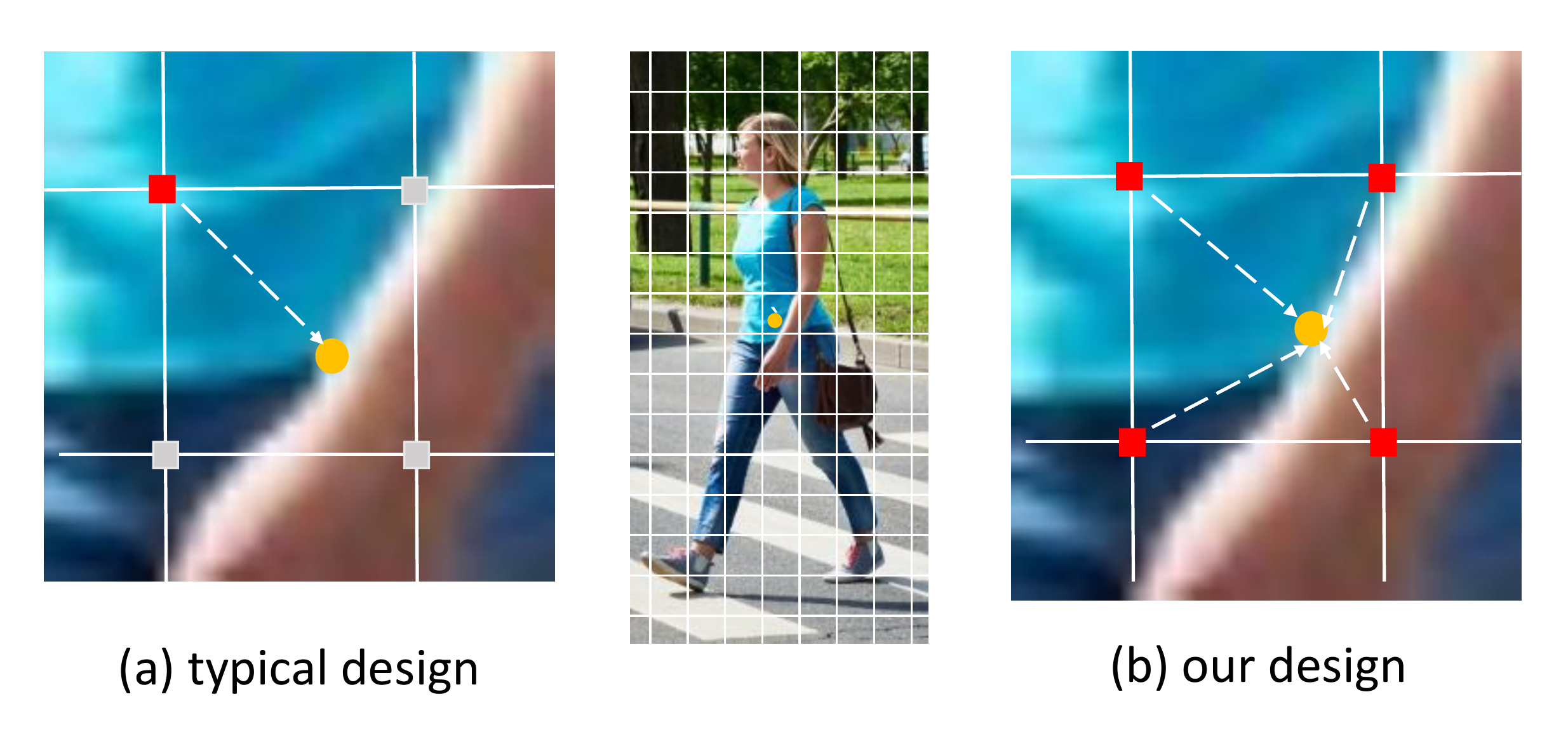}
	\end{center}
	%\vspace{-0.1in}
	\caption{Illustration for center and offset target. Circle point is the real center $(\frac{x_{k}}{r}, \frac{y_{k}}{r})$, while square boxes denote four points next to it. \color{red}Red \color{black}indicates positive, while \color{gray}grey \color{black} is negative. The dashed line denotes the offset, direction from positives to the real center.}
	\label{fig:newGT}
	%\vspace{-0.1in}
\end{figure}

\subsection{Optimization Target}

In the following, we present the optimization target for the conventional detection branches, including center, scale, and offset maps, which consists of the definition for ground truth targets as well as the objective functions.

\subsubsection{Center Loss}
The real location of an object's center point is $(\frac{x_{k}}{r}, \frac{y_{k}}{r})$. Instead of assigning a single position $(\lfloor \frac{x_{k}}{r} \rfloor, \lfloor \frac{y_{k}}{r} \rfloor)$ as positive in~\cite{liu2019high}, we assign all four corners next to the real location as positives while all others as negatives for the center target $\mathcal{C} \in \mathcal{R}^{\frac{W}{r} \times \frac{H}{r} \times 1}$. 

As in~\cite{liu2019high}, predicting the center map can be formulated as a classification task with the supervision of center target $\mathcal{C}$ via a penalty-reduced pixel-wise logistic regression with focal loss~\cite{lin2017focal} as below: 
\begin{equation}
L_{c} = -\frac{1}{K}\sum_{i=1}^{W/r}\sum_{j=1}^{H/r}\alpha_{ij}(1-\hat{p}_{ij})^{\gamma}log(\hat{p}_{ij}),\\
\label{eq:clsloss}
\end{equation}
where $K$ in Eq.~\ref{eq:clsloss} is the number of positive instances in an image. Let $p_{ij}\in[0,1]$ denote the predicted probability indicating whether the location $(i,j)$ is a positive instance or not. $\hat{p}_{ij}$ and $\alpha_{ij}$ are defined as follows:
\begin{equation}
\begin{aligned}
&\hat{p}_{ij}=
\begin{cases}
p_{ij}  & \mbox{if $y_{ij}=1$}\\
1-p_{ij}& \mbox{otherwise,}
\end{cases} \\
&\alpha_{ij}=
\begin{cases}
1  & \mbox{if $y_{ij}=1$}\\
(1-M_{ij})^{\beta}& \mbox{otherwise.}
\end{cases}
\end{aligned}
\end{equation}
where $y_{ij}=1$ is the ground truth label representing the positive location. To deal with the ambiguity from those negatives surrounding the positives, $\alpha_{ij}$ according to the Gaussian mask $M$ is applied to reduce their contributions to the total loss. As in~\cite{law2018cornernet}, Hyper-parameters $\gamma$  and $\beta$ are set to $2$ and $4$, respectively.

\subsubsection{Scale Loss}
As for the scale target $\mathcal{S} \in \mathcal{R}^{\frac{W}{r} \times \frac{H}{r} \times s}$, $s=1$ if only height is predicted. In CrowdHuman dataset, the aspect ratio is not fixed. $s$ is set to 2 so that both height and width can be predicted. For simplicity, we only consider the height prediction in the following illustration. we assign $\log(h_k)$ to the $k$-th positive regions, which is assigned to the negatives with an extension of one pixel to the positive region with the size of $4 \times 4$.

The prediction of scale map is formulated as a regression task with supervision $\mathcal{S}$ via SmoothL1 loss~\cite{girshick2015fast} as follows:
\begin{equation}
L_{s} = \frac{1}{K}\sum_{k=1}^{K}\Phi(\hat{s}_{k},s_{k}),
\label{eq:sml1}
\end{equation}
where $\Phi$ is SmoothL1 loss, $\hat{s}_{k}$, $s_{k}$ represent the predicted height and the ground truth of each positive instance, respectively. The supervision acts only at the positions that have been assigned with scale values. All other locations are ignored.

\subsubsection{Offset Loss}
For offset target $\mathcal{O} \in \mathcal{R}^{\frac{W}{r} \times \frac{H}{r} \times 2}$, we estimate the offset for each corner towards the real center in four directions, which means that offsets are $(\frac{x_{k}}{r}-\lfloor \frac{x_{k}}{r} \rfloor, \frac{y_{k}}{r}-\lfloor \frac{y_{k}}{r} \rfloor)$,  $(\frac{x_{k}}{r}-\lceil \frac{x_{k}}{r} \rceil, \frac{y_{k}}{r}-\lfloor \frac{y_{k}}{r} \rfloor)$,  $(\frac{x_{k}}{r}-\lfloor \frac{x_{k}}{r} \rfloor, \frac{y_{k}}{r}-\lceil \frac{y_{k}}{r} \rceil)$ and  $(\frac{x_{k}}{r}-\lceil \frac{x_{k}}{r} \rceil, \frac{y_{k}}{r}-\lceil \frac{y_{k}}{r} \rceil)$ for top-left, top-right, bottom-left and bottom-right point, respectively. Fig.~\ref{fig:newGT} illustrates our design.

The branch of offset map is also formulated as a regression task with the supervision of $\mathcal{O}$ via SmoothL1 loss as below:
\begin{equation}
L_{o} = \frac{1}{K}\sum_{k=1}^{K}\Phi(\hat{o}_{k},o_{k}),
\label{eq:sml1_off}
\end{equation}
where $\hat{o}_{k}$ and $o_{k}$ represent the predicted offset and the ground truth of each positive instance, respectively. Similar to the scale map, the supervision acts only at locations that have been assigned with offset values and all other locations are ignored.

The aforementioned ground truth targets enjoy several key advantages:
1) There is no apparent boundary between four corners. In contrast to the typical design, a detector may predict the other three corners as center as well. However, the offset direction is always bottom-right which makes the localization less precise. Our design regresses offsets in all four directions towards the real center making the localization more precise; 2) By treating a $2 \times 2$ region as positives, we increase the positive instances three times larger than the typical design that eases the extremely-class-imbalance issue during training to some extent; 3) Our design assigns more than one positive location to each ground truth bounding box, which enables us to design an effective attribute loss described in the following.

\subsection{Attribute Loss}
Besides the conventional loss for an anchor-free detector, we propose an attribute-aware loss to enforce the neural network to catch a high-level understanding of the crowd. The proposed loss consists of two different types of attribute losses, density loss $L_{den}$ and diversity loss $L_{div}$, with a weighted sum as below:
\begin{equation}
L_{a} = \lambda_{den} L_{den} + L_{div}
\end{equation}
where $\lambda_{den}$ is set to $5$ in our experiment. In this work, we aim to enable the neural networks to capture the high-level semantic information for a crowd. To this end, two attributes are proposed to describe a group of people, the density of people in a crowd and the diversity of a group. Given the predicted bounding box, the density attribute is the maximum IoU between the current bounding box and its surrounding ones. It gives hints to the proposed neural networks that a dense crowd might contain a number of people highly occluded from each other. Moreover, the diversity attribute is a metric for the algorithm to distinguish individuals among a crowd. With less diversity, a crowd should contain fewer people and the proposed detector outputs less bounding boxes. 

To achieve this, we generate a density target $\mathcal{D} \in \mathcal{R}^{\frac{W}{r} \times \frac{H}{r} \times 1}$ as the ground truth of density information.
Inspired by~\cite{liu2019adaptive}, for each position that the center of a ground truth box $b_i$ located on the $\mathcal{D}$, the value is defined as the maximum Intersection-over-Union~(\textrm{IoU}) between the ground truth bounding box $b_i$ and other nearby boxes in the image as follows
\begin{equation}
d_i \coloneqq \max\limits_{{b_j \in \mathcal{G}, i \neq j}} \textrm{IoU}(b_i, b_j), \label{eq:density}
\end{equation}
where $\mathcal{G}$ is the ground truth bounding box set of an image. 
As for diversity information, we implicitly encode it by proposing an attribute loss instead of designing a diversity ground truth target.
More specifically, both density and diversity information are encoded as an $m$-dimension vector $e \in \mathcal{R}^m$ embedded in each position of the attribute map.
With four locations labeled as the positive samples in center target, $N_p = 4$ embedding vectors $e_{k,i}, i=1,...,N_p$ are used to represent the attribute of an object.
We employ the norm of embedding vector as the density information. 
Estimation is formulated as a regression task via SmoothL1 loss as below
%\vspace{-0.1in}
\begin{equation}
L_{den} = \frac{1}{N}\sum^{N}_{k=1} \sum^{N_p}_{i=1} \Phi(\left\|e_{k,i}\right\|_2,d_{k})
\end{equation}
where $\left\|e_{k,i}\right\|_2$ is the norm of $e_{k,i}$, and $d_k$ is the ground truth density of $k$-th object. In this loss function, we aim at regressing the norm of $e_{k,i}$ to the density attribute $d_k$. The supervision happens only at positive locations and all other positions are ignored.

Let $e_{k,i}^*$ and $e_{k}^*$ denote the normalized embedding vectors and its mean from the $k$-th object respectively as follows
\begin{equation}
e_{k,i}^* = \frac{e_{k,i}}{\left\|e_{k,i}\right\|_2},
\end{equation}
\begin{equation}
e_{k}^* = \frac{1}{N_p} \sum^{N_p}_{k=1} e_{k,i}^*.
\end{equation}
Based on above notations, our proposed diversity loss can be formulated as below:
\begin{equation}
L_{div} = L_{1} + L_{2}
\end{equation}
\begin{equation}
    L_{1} = \sum^{N}_{k=1} 
    \sum^{N}\limits_{\substack{j=1 \\ j \neq k}}
    \max \left(0, \Delta - \left\| e_{k}^* - e_{j}^* \right\|_1 \right), 
    \label{eq:push_loss}
\end{equation}
\begin{equation}
    L_{2} = \frac{1}{N} \sum^{N}_{k=1} \sum^{N_p}_{i=1}
    \left\| e_{k,i}^* - e_{k}^* \right\|_2^2,
    \label{eq:pull_loss}
\end{equation}
The hyper-parameter $\Delta$ is set to $1$. The diversity of an object is defined from two perspectives, dispersion $L_{1}$ and gathering $L_{2}$. Dispersion means our embedding vectors of different objects should be apart in the vector space. $L_{1}$ loss punishes two vectors from the different identities $e_{k}^*$ and $e_{j}^*$ for being close. Meanwhile, gathering $L_{2}$ tries to attract embedding vectors that belong to the same identity since we have $N_p$ embedding vectors to represent an object.

\begin{figure}[!tb]
	\begin{center}
		\includegraphics[width=0.6\linewidth]{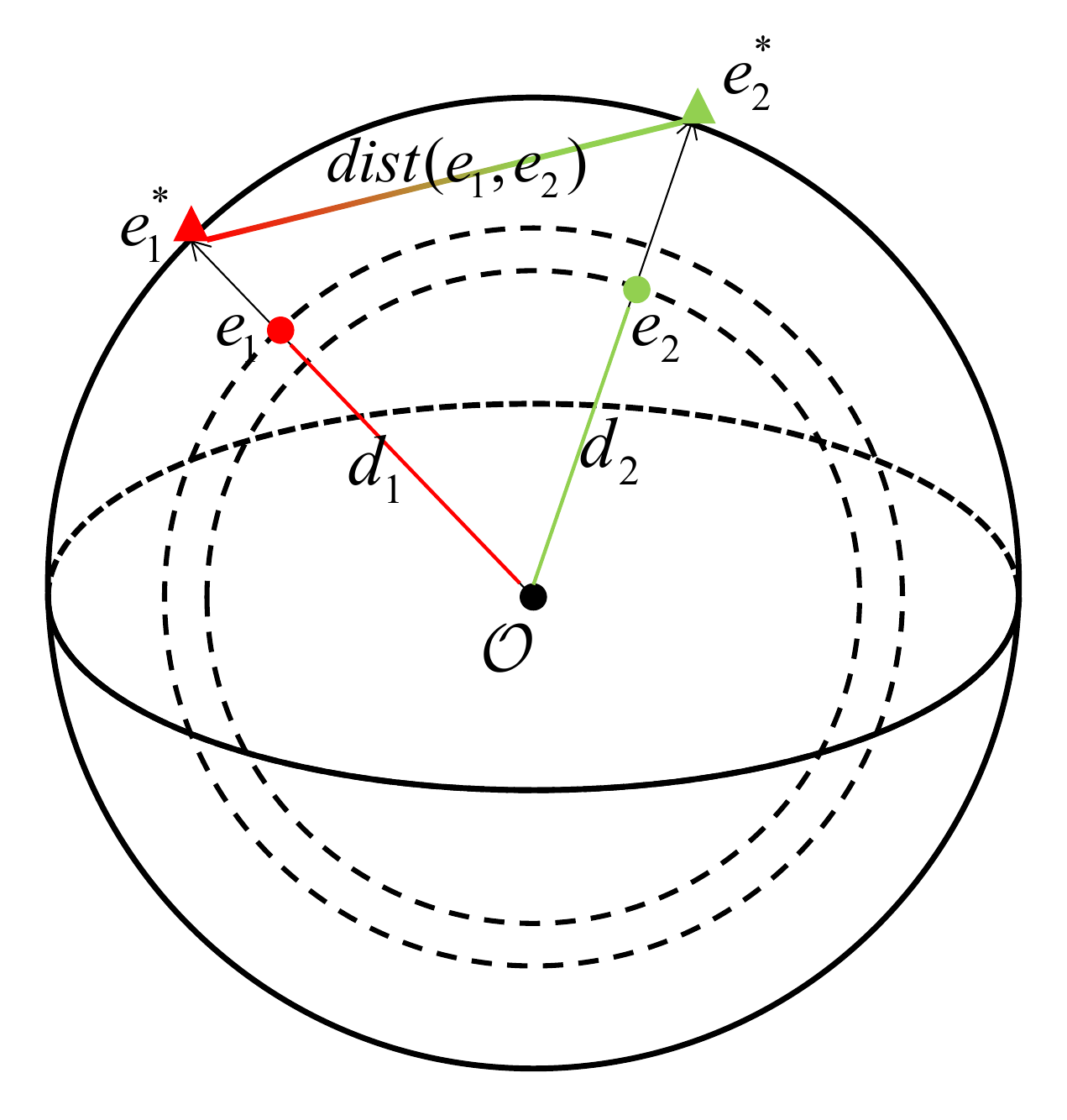}
	\end{center}
	% \vspace{-0.2in}
	\caption{A sphere $\mathcal{O}$ with radius $1$. $e_1$ and $e_2$ are embedding vectors and  $e_1^*$ and $e_2^*$ are the corresponding normalized embedding vectors. The length of red and green lines $d_1$ and $d_2$ denote the density while the changed color line denotes the distance of two embedding vectors. Vectors with similar density can still have large distance.}
	\label{fig:paradox}
	%\vspace{-0.1in}
\end{figure}
Note that a straightforward solution is to design two semantic feature maps. One is for the density map as in~\cite{liu2019adaptive} trained with $L_{den}$, and the other is for the diversity map trained with $L_{div}$, separately. However, we find that the overall network tends to be harder to optimize with more branches attached, which leads to deteriorating detection performance. Besides, a single branch of attribute map is much more elegant. As a matter of fact, we can not train an attribute map with dimension $m=1$ as in~\cite{liu2019adaptive} to encode both density and diversity information. The paradox lies in that, there are two predicted bounding boxes $b_1$, $b_2$ having the largest \textrm{IoU} with each other, denoted as $d$. Our objective function targets density prediction at value $d$ for both $b_1$ and $b_2$, thus with a metric distance $0$. However, the diversity prediction requires $b_1$, $b_2$ to have a large metric distance.

We show that with embedding vector dimension $m(>1)$ of an attribute map, these two properties can be held simultaneously. Taking an embedding space of dimension $m=3$ as an example, the embedding vector will fall in the sphere of radius $1$ since the density is a value between $0$ and $1$. 
In our design, the density is defined as the length of vector while the diversity is implicitly trained by restricting Euclidean distance between the normalized embedding vectors.
Density loss $L_{den}$ is defined on the length of each vector ($e_1$, $e_2$) while diversity loss $L_{div}$ is defined on the distance between the paired vectors. 
As shown in Fig. \ref{fig:paradox}, two embedding vectors with similar density, $e_1$ and $e_2$ can still have a large distance, $dist(e_1, e_2)$. In our experiment, we set $m=4$ if not specified.

Finally, we give a brief proof that larger Euclidean distance between normalized embedding vectors $e_1^*$, $e_2^*$ is equivalent to larger angle between embedding vectors $e_1$, $e_2$, and vice versa.
\begin{equation}
\begin{split}
\left\| e_{k}^* - e_{j}^* \right\|_2^2 &= \left\| e_{k}^*\right\|_2^2 + \left\| e_{j}^*\right\|_2^2 - 2\left\| e_{k}^*\right\|_2\left\| e_{j}^*\right\|_2\cos(\theta)\\
&= 2 - 2\cos(\theta),
\end{split}
\end{equation}
where $\theta=\left<e_{k}, e_{j}\right>$ is the angle between $e_{k}$ and $e_{j}$. In a geometrical view, our diversity loss actually restricts the angle between embedding vectors. In summary, training with our attribute loss $L_{a}$, the length $d$ of embedding vector can denote the density of each positive instance while the angle $\theta$ between two vectors can indicate the similarity of two predicted boxes.

The overall objective function can be derived as follows
\begin{equation}
L = \lambda_{c} L_{c} + \lambda_{s} L_{s} +\lambda_{o} L_{o} +\lambda_{a} L_{a},
\label{eq:jointloss}
\end{equation}
where $\lambda_{c}$, $\lambda_{s}$, $\lambda_{o}$ and $\lambda_{a}$ are empirically set to 0.01, 1, 0.03 and 0.01, respectively.

\subsection{Attribute-aware NMS Algorithm}
NMS is a critical post-processing step for object detectors. 
In this section, we first revisit greedy-NMS, and then introduce a density-aware NMS~\cite{liu2019adaptive}. Moreover, we propose a diversity-aware NMS algorithm to utilize the encoded diversity information from attribute map. To fully exploit our presented attribute map, we propose an attribute-aware NMS algorithm to refine the predicted bounding boxes more effectively by incorporating both density and diversity information.
\subsubsection{Greedy-NMS}
Given a set of predicted bounding boxes $\mathcal{B}$ with their corresponding confidence scores $\mathcal{S}$, greedy-NMS firstly selects the one $\mathcal{M} \in \mathcal{B}$ with the maximum score and all boxes in $\mathcal{B}$ which have the overlap with $\mathcal{M}$ larger than the threshold $N_t$ are removed. The algorithm continues to select the next bounding box in the remaining set of $\mathcal{B}$ and repeats the process until the end. The suppressing step is as follows:
\[
s_i=
\begin{cases}
s_i,& \textrm{IoU}(\mathcal{M}, b_i) < N_t\\
0,              &  \textrm{IoU}(\mathcal{M}, b_i) \ge N_t
\end{cases}
\;,
\]
where $\mathcal{M}$ is the bounding box with highest score in $\mathcal{B}$ and $s_i$ is the confidence score of a box $b_i$. $\textrm{IoU}$ represents the intersection over union score. $N_t$ is a constant threshold that is typically set to $0.5$. 
\begin{figure}[!t]
	\begin{center}
		\includegraphics[width=1\linewidth]{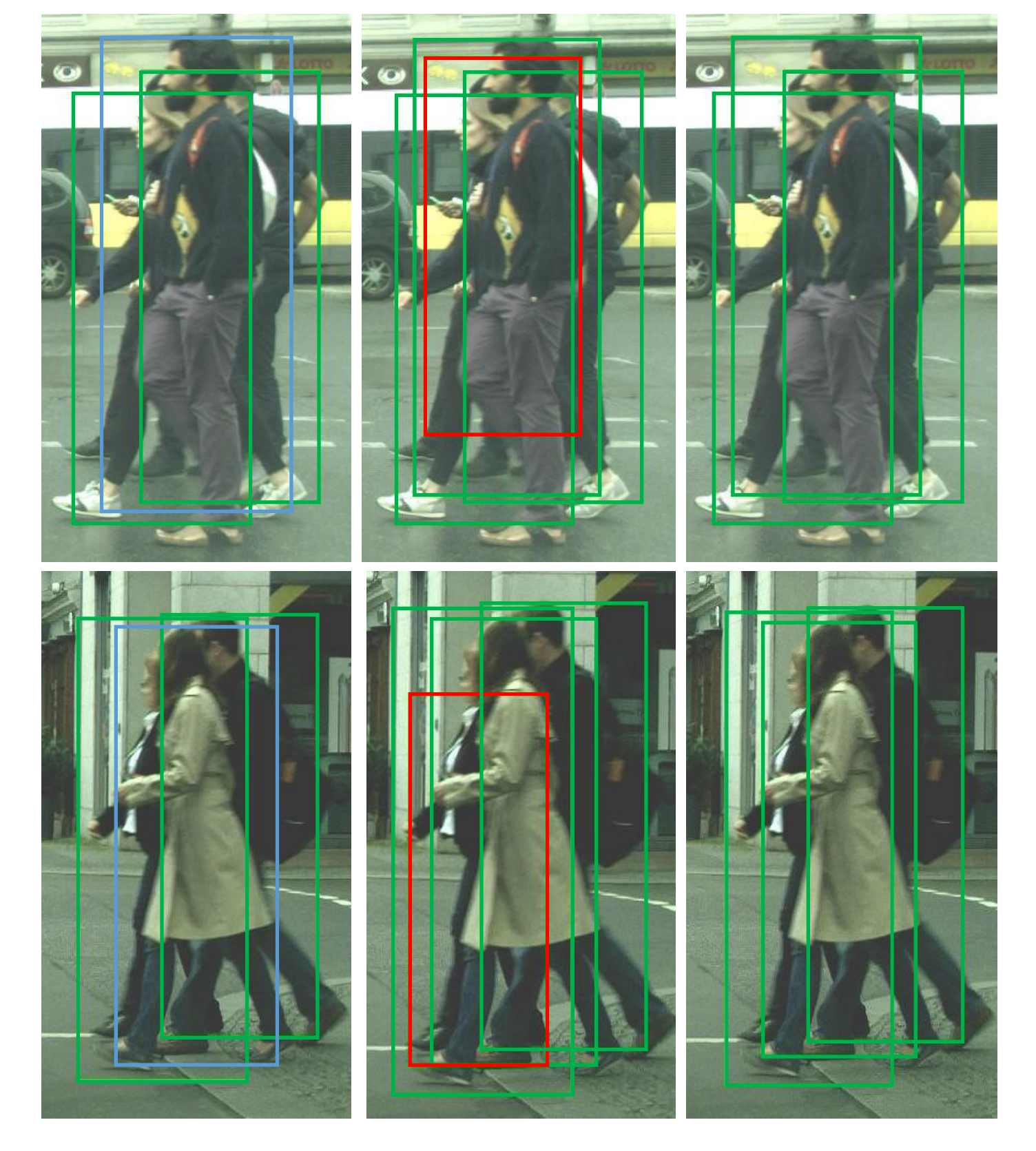}
	\end{center}
	%\vspace{-0.1in}
	\caption{Illustrative examples for different NMS algorithm. There are three pedestrians in each picture. The first column is the result of greedy-NMS, the mid column is density-aware NMS, and the last column is our attribute-aware NMS algorithm. The missed bounding box is colored by blue, and the false positive is denoted as red. There are three pedestrians in each picture, greedy
	}
	\label{fig:whyIA}
	%\vspace{-0.1in}
\end{figure}

\subsubsection{Density-aware NMS}

Similar to~\cite{liu2019adaptive}, the density-aware NMS is designed to incorporate the density information, which is able to adaptively adjust
the threshold $N_t$ by the density of each bounding box.
The suppressing step of density-aware NMS can be described as below
\begin{equation} \label{eq:ada}
N_{\mathcal{M}}:=\max(N_t, d_{\mathcal{M}}),
\end{equation}
\[
s_i=
\begin{cases}
s_i,& \textrm{IoU}(\mathcal{M}, b_i) < N_{\mathcal{M}}\\
0, &  \textrm{IoU}(\mathcal{M}, b_i) \ge N_{\mathcal{M}}
\end{cases}
\;,
\]
where $d_{\mathcal{M}}$ is the density of bounding box $\mathcal{M}$ defined in Eq.~\ref{eq:density}. $N_{\mathcal{M}}$ denotes the threshold for $\mathcal{M}$. 

We argue that the purpose of NMS algorithm is to suppress the predicted bounding boxes that belong to the same identity, which preserves the predicted bounding boxes from different identities.
However, neither greedy-NMS nor density-aware NMS addresses these two problems at the same time. While greedy-NMS sets the threshold blindly, density-aware NMS may work well in keeping predicted boxes in crowd scenes, however, it also preserves more boxes belong to the same identity thus leading to more false positives. An illustrative example is shown in Fig.~\ref{fig:whyIA}.

\subsubsection{Attribute-aware NMS and Diversity-aware NMS}

\begin{figure}[t]
	\begin{center}
		\includegraphics[width=1.0\linewidth]{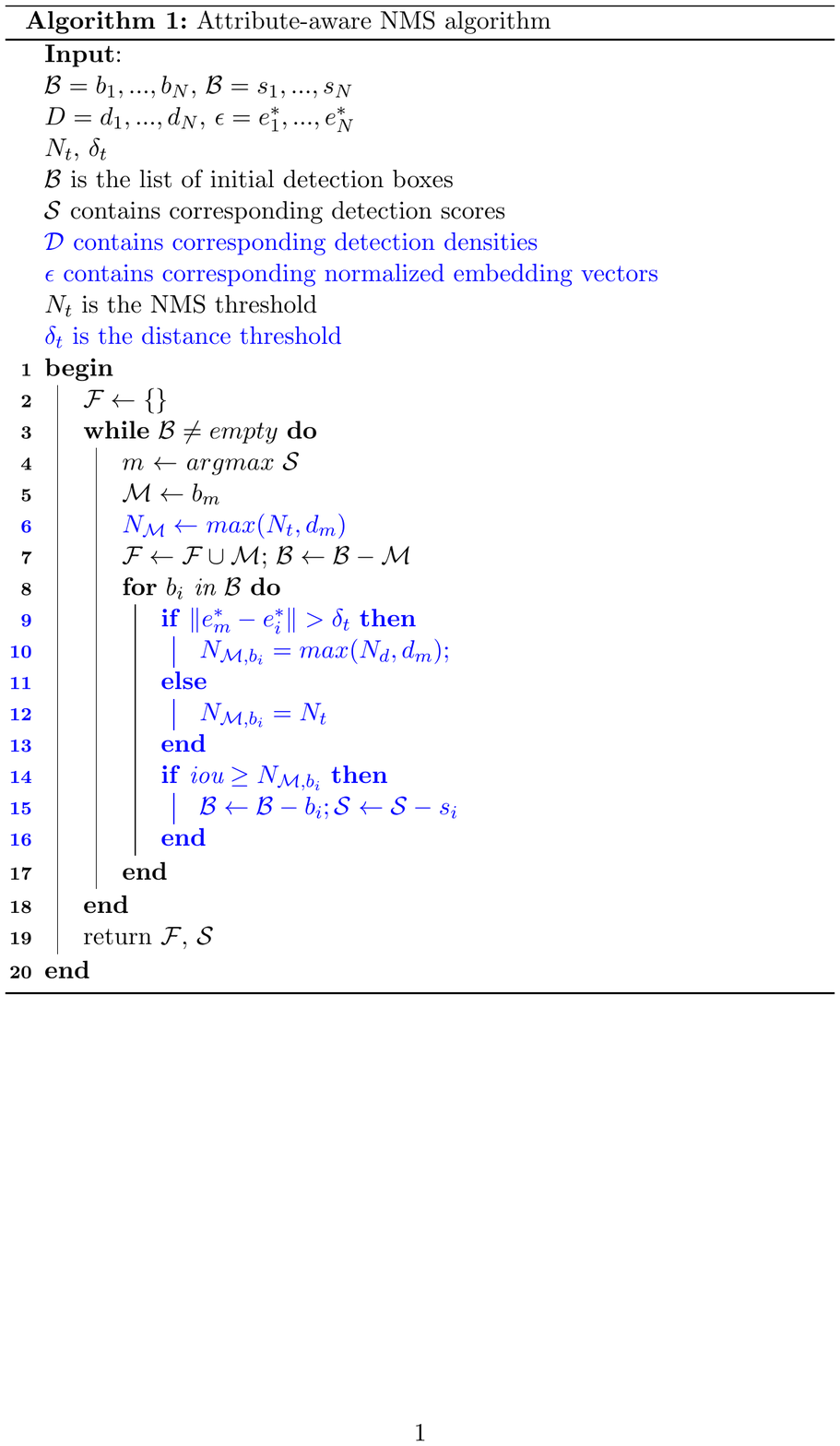}
	\end{center}
	%\vspace{-0.3in}
	\caption{The psudo code for our attribute-aware NMS. The difference from greedy-NMS is highlighted as blue.}
	\label{fig:algorithm}
	%\vspace{-0.2in}
\end{figure}
To incorporate diversity information, we present a diversity-aware NMS algorithm which takes the similarity between two compared bounding boxes into consideration.
The suppressing step of diversity-aware NMS is defined as below
\begin{equation}
N_{\mathcal{M}, b_i}:=
\begin{cases}
N^{high}_t,& dist(\mathcal{M}, b_i) > \delta_t\\
N^{low}_t, &  dist(\mathcal{M}, b_i) \le \delta_t
\end{cases}
\;,
\end{equation}
\[
s_i=
\begin{cases}
s_i,& \textrm{IoU}(\mathcal{M}, b_i) < N_{\mathcal{M}, b_i}\\
0,              &  \textrm{IoU}(\mathcal{M}, b_i) \ge N_{\mathcal{M}, b_i}
\end{cases}
\;,
\]
$N^{high}_t, N^{low}_t$ are constant thresholds, which are empirically set to $0.6$ and $0.5$. $dist(*)$ denotes a distance function to measure the diversity between two bounding boxes. The distance threshold $\delta_{t}$ is set to $0.9$. Diversity-aware NMS also has trouble setting distance threshold and constant $N^{high}_t$ in real complicated crowd scenarios.

To fully exploit our attribute map, we propose a novel post-processing algorithm called attribute-aware NMS by taking both density and diversity information of each box into account.
The suppressing step of attribute-aware NMS is defined as follows
\begin{equation} \label{eq:aed}
N_{\mathcal{M}, b_i}:=f(d_{\mathcal{M}}, dist(\mathcal{M}, b_i); N_t, \delta_t),
\end{equation}
\[
f(d, dist; N_t, \delta_t)=
\begin{cases}
\max(d, N_t),& dist > \delta_t\\
N_t, &  dist \le \delta_t
\end{cases}
\;,
\]
\[
s_i=
\begin{cases}
s_i,& \textrm{IoU}(\mathcal{M}, b_i) < N_{\mathcal{M}, b_i}\\
0, &  \textrm{IoU}(\mathcal{M}, b_i) \ge N_{\mathcal{M}, b_i}
\end{cases}
\;,
\]
where $d_{\mathcal{M}}$ represents the density of box $\mathcal{M}$. $dist(*)$ is a distance function that accounts for the diversity information between two predicted boxes. Note that the threshold $N_{\mathcal{M}, b_i}$ is not only a function of $\mathcal{M}$ but also related to the box $b_i$. We discuss the properties of our designed threshold function $N_{\mathcal{M}, b_i}$ and suppression strategy in the following.
\begin{itemize}
	\item $N_{\mathcal{M}, b_i}$ is high if and only if when $d_{\mathcal{M}}$ is large ($\mathcal{M}$ has a high density) and $dist(\mathcal{M}, b_i)$ is large (the two compared boxes $\mathcal{M}$ and $b_i$ belong to different identities).
	\item When $\mathcal{M}$ is not in a crowd scene, $N_{\mathcal{M}, b_i}$ will be low since $d_{\mathcal{M}}$ is small.
	\item When $\mathcal{M}$ locates in the crowded region ($d_{\mathcal{M}}$ is high) and $\mathcal{M}$, $b_i$ belong to the same identity ($dist(\mathcal{M}, b_i)$ is small), threshold $N_{\mathcal{M}, b_i}$ will be decreased to suppress $b_i$ thus reducing false positives.
	\item When $\mathcal{M}$ locates in the crowded region and and $\mathcal{M}$ and $b_i$ belong to different identities ($dist(\mathcal{M}, b_i)$ is large), $N_{\mathcal{M}, b_i}$ will be high thus neighboring boxes which belong to different identities are preserved.
\end{itemize}
Thus, our design satisfies the two conditions mentioned before simultaneously. As shown in Fig.~\ref{fig:whyIA}c, our attribute-aware NMS can address the issue in a crowd. The algorithm is formally described in Fig~\ref{fig:algorithm}. During inference, the predicted density and $dist(*)$ function is calculated as follows
\begin{equation}
dist(b_i, b_j) = \left\| e_{i}^* - e_{j}^* \right\|_2
\end{equation}
\begin{equation}
d_i = \left\| e_{i} \right\|_2
\end{equation}
where $e_{i}$ and $e_{i}^*$ denotes the embedding vector and the normalized embedding vector of the predicted box $b_i$, respectively.
%\vspace{-0.1in}
\section{Experiments} %\label{exp}
In this section, we first introduce our experimental settings and implementation details. Then, we conduct an ablative analysis for each component of our proposed pedestrian detector on Citypersons dataset~\cite{DBLP:conf/cvpr/ZhangBS17}. Furthermore, we compare our approach with the state-of-the-art methods on both Citypersons and CrowdHuman datasets~\cite{shao2018crowdhuman}. 

{
\setlength{\tabcolsep}{18pt}
\begin{table*}[!htb]
	\begin{center}
		% \resizebox{.9\textwidth}{!}{
		\begin{tabular}{l|c|c|c c c|c}
			\hline
			Method & Backbone & Reasonable & Heavy & Partial & Bare & Test Time\\
			\hline
			\hline
			FRCNN~\cite{DBLP:conf/cvpr/ZhangBS17} & VGG-16 & 15.4 & - & - & - & -\\
			\hline
			FRCNN+Seg~\cite{DBLP:conf/cvpr/ZhangBS17} & VGG-16 & 14.8 & - & - & - & -\\
			\hline
			OR-CNN~\cite{zhang2018occlusion} & VGG-16 & 12.8 & 55.7 & 15.3 & 6.7 & -\\
			\hline
			RepLoss~\cite{wang2018repulsion} & ResNet-50 & 13.2 & 56.9 & 16.8 & 7.6 & -\\
			\hline
			TLL~\cite{song2018small} & ResNet-50 & 15.5 & 53.6 & 17.2 & 10.0 & -\\
			\hline
			TLL+MRF~\cite{song2018small} & ResNet-50 & 14.4 & 52.0 & 15.9 & 9.2 & -\\
			\hline
			ALFNet~\cite{liu2018learning} & ResNet-50 & 12.0 & 51.9 & 11.4 & 8.4 & 0.27s/img\\
			\hline
			WIDERPERSON~\cite{7792742} & VGG-16 & {11.1} & {-} & {-} & {-} & -\\
			\hline
			CSP~\cite{liu2019high} & ResNet-50 & {11.0} & {49.3} & {10.4} & {7.3} & 0.33s/img\\
			\hline
			Adaptive-NMS~\cite{liu2019adaptive} & ResNet-50 & {10.8} & {54.0} & {11.4} & {6.2} & -\\
			\hline
			APD (ours) & ResNet-50 &{10.6} & {49.8} & {9.5} & {7.1} & 0.12s/img\\
			\hline
			APD (ours) & DLA-34 & \textbf{8.8} & \textbf{46.6} & \textbf{8.3} & \textbf{5.8} & 0.16s/img\\
			\hline
		\end{tabular}
		% }
	\end{center}
	\caption{Comparison with state-of-the-art methods on CityPersons validation set. All models are trained on the training set. Our result is tested on the original image size $1024 \times 2048$. The reported testing time is under the same environment with the input size of $1024 \times 2048$ on Nvidia GTX 1080Ti GPU.}
	\label{table:cityval}
\end{table*}
}

%\vspace{-0.1in}
\subsection{Experiment Setup}
To evaluate the efficacy of our proposed pedestrian detector, we conduct experiments on the challenging benchmarks in a crowd, including Citypersons and CrowdHuman.

\textbf{Datasets}
CityPersons~\cite{DBLP:conf/cvpr/ZhangBS17} is a newly collected pedestrian detection dataset built on top of the CityScapes benchmark~\cite{Cordts2016Cityscapes}, which includes 5,000 images (2,975 for training, 500 for validation and 1,525 for testing). It is a challenging dataset with various occlusion levels. We train our model on the official training set with 2975 images and test on the validation set with 500 images and test set with 1525 images. CrowdHuman dataset~\cite{shao2018crowdhuman} has recently been released to specifically target to the crowd issue in the human detection task, which has 15,000, 4370, and 5,000 images from the Internet for training, validation, and test, respectively. It is of much higher crowdedness than CityPersons. We train on the training set and test on the validation set, where the full body region annotations are used for training and evaluation. In all the two datasets, we employ log-average Miss Rate over False Positive Per Image (FPPI) as the evaluation metric, which ranges in $[10^{-2}, 10^{0}]$ (denoted as $MR^{-2}$).

\textbf{Training Details}
We implement our proposed method in Pytorch~\cite{paszke2017automatic}.
Adam~\cite{kingma2014adam} optimizer is adopted to optimize the network. Also, moving average weights~\cite{tarvainen2017mean} is used to achieve better and more stable training. The input sizes of training images are $640 \times 1280$ and $768 \times 1152$ for CityPersons and CrowdHumans, respectively. While resizing the inputs size, we keep the original aspect ratio for input images. For CityPersons, we optimize the network on two GPUs (GTX 1080Ti) with four images per GPU for a mini-batch. The learning rate is set to $1\times10^{-4}$. Training process  is stopped after $37.5K$ iterations. For CrowdHuman, a mini-batch contains 16 images with four GPUs (GTX 1080Ti), and the learning rate is set to $1\times10^{-4}$. Training process is stopped after $100K$ iterations.

\subsection{Ablation Study on CityPersons}
We conduct an ablative analysis of the proposed method on CityPersons dataset. To this end, each component of our proposed detector is evaluated, including our new ground truth target design, our attribute map as well as attribute-aware NMS algorithm and the sensitivity of our embedding dimension $m$. The original image size $1024 \times 2048$ is used for testing.

\begin{table}[!htb]
	\begin{center}
		\begin{tabular}{l|c|c|c c c}
			\hline
			Backbone & GT Targets & Reasonable & Heavy & Partial & Bare \\
			\hline
			\hline
			ResNet-50 & & 11.2 & 50.0 & 10.2 & 7.8 \\
			\hline
			ResNet-50 & \checkmark & 11.0 & 49.8 & 10.1 & 7.4 \\
			\hline
			\hline
			DLA-34 & & 10.2 & 48.1 & 9.6 & 6.8 \\
			\hline
			DLA-34 & \checkmark & 9.5 & 47.4 & 8.6 & 6.6 \\
			\hline
		\end{tabular}
	\end{center}
	\caption{Comparison to typical ground truth design. GT Targets imply our new design of ground truth targets. Our design performs consistently better than typical design in both ResNet-50 and DLA-34 architecture.}
	\label{table:abla_gt_maps}
\end{table}

\subsubsection{Ground truth} We compare our ground truth targets design with typical configuration. Our design outperforms conventional method consistently in both ResNet-50 and DLA-34 architecture. It outperforms $0.2\%$ in ResNet-50. Moreover, it increases to $0.7\%$ with a more powerful backbone DLA-34. The performance gain is because our center target design assigns more locations to the positives which eases the extremely class imbalance issue during training, while the offset target design makes the localization more precise. Results are reported in Table \ref{table:abla_gt_maps}. If not specified, we use our new design in the following experiments.

\begin{table}[!htb]
	\begin{center}
		\begin{tabular}{l|c|c|c c c}
			\hline
			Backbone & Attribute & Reasonable & Heavy & Partial & Bare \\
			\hline
			\hline
			ResNet-50 & & 11.0 & 49.8 & 10.1 & 7.4 \\
			\hline
			ResNet-50 & \checkmark &{10.6} & {49.8} & {9.5} & {7.1} \\
			\hline
			\hline
			DLA-34 & & {9.5} & {47.4} & {8.6} & {6.6} \\
			\hline
			DLA-34 & \checkmark & {8.8} & {46.6} & {8.3} & {5.8} \\
			\hline
		\end{tabular}
	\end{center}
	\caption{Comparison of our attribute-aware pedestrian detector with non-attribute version. Our detector performs consistently better in both ResNet-50 and DLA-34 architecture.}
	\label{table:abla_attr}
\end{table}
{
\setlength{\tabcolsep}{2.5pt}
\begin{table}[!htb]
	\begin{center}
		\begin{tabular}{l|c|c|c|c|c c c}
			\hline
			\multirow{2}*{Backbone} & \multicolumn{2}{c|}{Separate Branches} & \multirow{2}*{Attribute} & \multirow{2}*{Reasonable} & \multirow{2}*{Heavy} & \multirow{2}*{Partial} & \multirow{2}*{Bare} \\
			\cline{2-3}
			 & Density & Diversity & & & & & \\
			\hline
			\hline
			DLA-34 & & & &{9.5} & {47.4} & {8.6} & {6.6} \\
			\hline
			DLA-34 & \checkmark & & & {9.2} & {46.5} & {8.7} & {5.9} \\
			\hline
			DLA-34 & & \checkmark & & {9.3} & {47.8} & {8.6} & {6.1} \\
			\hline
			DLA-34 & \checkmark & \checkmark & & {9.2} & {47.7} & {8.9} & {5.9} \\
			\hline
			DLA-34 & & & \checkmark & {8.8} & {46.6} & {8.3} & {5.8} \\
			\hline
		\end{tabular}
	\end{center}
	\caption{Comparison of our attribute-aware pedestrian detector with solely density prediction or diversity prediction or two separate branches for density and diversity prediction. Our attribute-aware pedestrian detector performs better than only density prediction or diversity prediction. Our attribute-aware NMS algorithm still takes effects when predicting density and diversity in two separate branches. However, more branches lead to deteriorate the overall detection performance.}
	\label{table:abla_attr2}
\end{table}
}

\subsubsection{Attribute map v.s. attribute-aware NMS}
We first evaluate the effectiveness of our attribute-aware pedestrian detector with both ResNet-50 and DLA-34 architecture. As shown in Fig.~\ref{table:abla_attr}, our detector outperforms $0.4\%$ and $0.7\%$ in ResNet-50 and DLA-34 architecture, respectively. We further conduct an experiment to compare our attribute-aware pedestrian detector with the following three settings: 1) a density map that predicts density information together with density-aware NMS which follows the strategy of adaptive-NMS~\cite{liu2019adaptive}; 2) a diversity map that predicts diversity information together with diversity-aware NMS,
and two separate branches, density map and diversity map, together with attribute-aware NMS.

As shown in Table~\ref{table:abla_attr2}, our attribute-aware pedestrian detector outperforms $0.7\%$ than the default setting with greedy-NMS. Compared with density prediction with density-aware NMS or diversity prediction with diversity-aware NMS, our detector still outperforms $0.4\%$ and $0.5\%$, respectively. As for predicting density and diversity in two separate branches with attribute-aware NMS, our NMS algorithm still outperforms $0.3\%$ than the default setting with greedy-NMS. However, our attribute map that encodes both density and diversity information simultaneously outperforms $0.4\%$ than two separate branches, which we argue that more detection heads deteriorate the detection performance, and this validates our superior design for attribute map.
{
\setlength{\tabcolsep}{8pt}
\begin{table}[!htb]
	\begin{center}
			\begin{tabular}{l|c|c|c c c}
				\hline
				Backbone & $m$       & Reasonable & Heavy & Partial & Bare \\
				\hline
				\hline
				DLA-34 & 2 & 9.2 & 46.6 & 8.6 & 6.1 \\
				\hline
				DLA-34 & 3 & 9.0 & 47.0 & 8.2 & 5.9 \\
				\hline
				DLA-34 & 4 & 8.8 & 46.6 & 8.3 & 5.8 \\
				\hline
				DLA-34 & 8 & 9.0 & 47.1 & 8.4 & 5.9 \\
				\hline
			\end{tabular}
	\end{center}
	\caption{Sensitivity of attribute-aware pedestrian detector to $m$. Embedding length $m=4$ yields the best result.}
	\label{table:abla_m}
\end{table}
}

\subsubsection{Dimension $m$ of embedding vector}
We analyze the sensitivity of our attribute-aware pedestrian detector to the dimension $m$ of embedding vector. We conduct experiments by varying $m$ with $2$, $3$, $4$, and $8$. The best result is obtained when $m=$  $4$. However, the results are comparable for $3$, $4$ or $8$, as shown in Table~\ref{table:abla_m}. 

\subsection{Results on CityPersons}
The proposed approach is extensively compared with state-of-the-art methods. Our presented attribute-aware pedestrian detector outperforms all previous methods at a large margin on both validation and the test set. Specifically, we achieve $MR^{-2}$ of $8.8\%$ on the Reasonable setting in validation set~\ref{table:cityval}, which is $2\%$ better than the best competitor($10.8\%$ of Adaptive-NMS). Our method also performs consistently better in all other three settings, $46.6\%$, $8.3\%$ and $5.8\%$ in Heavy, Partial and Bare, respectively. As shown in Table~\ref{table:citytest}, we achieve $MR^{-2}$ of $8.27\%$ on the Reasonable setting in test set, which is $3.05\%$ better than the best competitor($11.32\%$ of OR-CNN). Our detector also outperforms consistently in Small, Heavy and All settings, $3.16\%$, $15.98\%$, $4.54\%$, respectively. Furthermore, the proposed approach runs faster than recent state-of-the-art method, which achieves $0.16s/img$ compared with $0.27s/img$ of ALF~\cite{liu2018learning} and $0.33s/img$ of CSP~\cite{liu2019high}.
\begin{table}[!htb]
	\begin{center}
		% \resizebox{.5\textwidth}{!}{
		\begin{tabular}{l|c|c|c c c}
			\hline
			Method & Backbone & Reasonable & Small & Heavy & All \\
			\hline
			\hline
			FRCNN~\cite{DBLP:conf/cvpr/ZhangBS17} & VGG-16 & 12.97 & 37.24 & 50.47 & 43.86\\
			\hline
			OR-CNN~\cite{zhang2018occlusion} & VGG-16 & 11.32 & 14.19 & 51.43 & 40.19 \\
			\hline
			RepLoss~\cite{wang2018repulsion} & ResNet-50 & 11.48 & 15.67 & 52.59 & 39.17 \\
			\hline
			APD (ours) & DLA-34 & \textbf{8.27} & \textbf{11.03} & \textbf{35.45} & \textbf{35.65} \\
						\hline
		\end{tabular}
		% }
	\end{center}
	\caption{Comparisons with the state-of-the-art methods on CityPersons testing set. All models are trained on the training set. Our result is tested with the original image size of $1024 \times 2048$.}
	\label{table:citytest}
\end{table}

\subsection{Results on CrowdHuman}
We compared our proposed detector with the most recent state-of-the-art methods, including Adaptive-NMS~\cite{liu2019adaptive} and FRCNN+FPN. In~\cite{liu2019adaptive}, the result of both single-stage and two-stage anchor-based methods are provided. As shown in Table~\ref{table:crowdhuman}, the proposed pedestrian detector with our new ground truth targets performs better than the competitor at a clear margin. Even compared with two-stage FPN, we still outperform a lot $36.79$ compared with $49.73$. Using the attribute-aware NMS algorithm, we can further improve our result from $36.79$ to $35.76$. It demonstrates the efficacy of our proposed detector.

{
\setlength{\tabcolsep}{15pt}
\begin{table}[!htb]
	\begin{center}
		%\resizebox{0.5\textwidth}{!}{
			\begin{tabular}{l|c}
				\hline
				Method       & $MR^{-2}$ \\
				\hline
				\hline
				FRCNN+FPN~\cite{shao2018crowdhuman} & 50.42  \\
				RetinaNet~\cite{shao2018crowdhuman} & 63.33  \\
				FPN+adaptive-NMS~\cite{liu2019adaptive} & 49.73     \\
				RFB+adaptive-NMS~\cite{liu2019adaptive} & 63.03      \\
				\hline
				APD w/o (ours) & 36.79     \\
				APD (ours) & \textbf{35.76}	  \\
				\hline
			\end{tabular}
		%}
	\end{center}
	\caption{Evaluation of full body detections on the CrowdHuman validation set. w/o denotes that attribute-aware NMS is not used and only our new ground truth targets are adopted.}
	\label{table:crowdhuman}
\end{table}
}

\section{Conclusions}
In this paper, we proposed an attribute-aware pedestrian detector to explicitly capture the semantic attributes in a high-level feature detection manner. A pedestrian-oriented attribute map was employed to encode both density and diversity information of each identity simultaneously.
To distinguish the pedestrian from a highly overlapped group, a novel attribute-aware Non-Maximum Suppression~(NMS) was proposed to adaptively reject the false-positive results in a very crowd settings. Moreover, we designed the novel ground truth targets to alleviate the training difficulties caused by the attribute configuration and extremely class imbalance. Finally, we conducted the extensive experiments to demonstrate the efficacy of our proposed pedestrian detector. As a result, our approach outperformed the state-of-the-art methods at a large margin on both CityPersons and CrowdHuman datasets. For our future work, we plan to extend our approach for instance segmentation in crowd.

\ifCLASSOPTIONcaptionsoff
  \newpage
\fi


% \clearpage
% \newpage
\bibliographystyle{abbrv}
\bibliography{ref}
\end{document}